\documentclass{acmsiggraph}

\usepackage{amsmath}
\usepackage{graphicx}

\title{Large Angle based Skeleton Extraction for 3D Animation}

\author{Hugo Martin, Rapha{\"e}l Fernandez, Yong Khoo}
\pdfauthor{Hugo Martin}


\setcopyright{rightsretained}

\copyrightyear{2016}

\begin{document}

\maketitle

\begin{abstract}
 In this paper, we present a solution for arbitrary 3D character deformation by investigating  rotation angle of decomposition and preserving the mesh topology structure.  In computer graphics,  skeleton extraction and skeleton-driven animation is an active areas and gains increasing interests from researchers. The accuracy is critical for realistic animation and related applications. There have been extensive studies on skeleton based 3D deformation. However for the scenarios of large angle rotation of different body parts, it has been relatively less addressed by the state-of-the-art, which often yield unsatisfactory results. Besides 3D animation problems, we also notice for many 3D skeleton detection or tracking applications from a video or depth streams, large angle rotation is also a critical factor in the regression accuracy and robustness. We introduced a distortion metric function to quantify the surface curviness before and after deformation, which is a major clue for large angle rotation detection. The intensive experimental results show that our method is suitable for 3D modeling, animation,  skeleton based tracking applications.
\end{abstract}

\keywordlist

\conceptlist

\printcopyright

\section{Introduction}
Recently, it has been proposed several technology-based optimization scheme, the task \cite{1}\cite{3}. Most of these priorities is to get a new attitude, a few points deformation process a given character in the displacement of the guide is connected to the model. This type of interaction, you can drag the mouse on a two-dimensional canvas window, which provides a very intuitive interface to create animations. There is no limit mouse drag, that is, restrictions may be isolated or stretching may be required to rotate freely and not explicitly specified. However, it is a model of the desired shape of the idea should be retained as far as possible to the user as much as possible, he is manipulating a real object. Morphing technology to avoid any unnatural cut and scaled model is of particular interest, because the details of the global and local shape retention after editing \cite{17}\cite{18}\cite{2}.  With this paradigm for interactive image distortion. In this work, the key problem is how to find the optimal rigid transformation rotational component effective \cite{15}. They proposed a closed form solution that uses a similar relationship between conversion and rigidity to obtain attractive results. Used as a two-dimensional image input and a set of control points of the algorithm. The user drags the control points, the calculated change in image conversion limit is "rigid" Every element of the image (not necessarily pixels). Calculate different solutions for each image element they use mobile Least Squares Method \cite{4}.

In  it uses the method described Gaussian process dimensionality reduction annealing track and articulated body movements classified particle filter. "This is a Gaussian process annealing particle filter \cite{Nie06}\cite{5}. Robustness score annealing particle filter layer and fewer errors, failed to embrace the movement and cross-category classification motion blur type of authentication required in this system \cite{6}. The method "described in the real-time three-dimensional non-parametric belief propagation fast track hinge mechanism. "Recursive Bayesian tracking articulated objects technology. In this system, good results are displayed in a variety of arm movements to track the location of people and slow processing rates. The method described in the "real-time three-axis inertial body using articulation / magnetic sensor package tracking." Kalman algorithm to use fusion. The algorithm further dynamic direction and position of the body portion is applied to obtain. Application development system and algorithm, the arm exercise test. The system uses three-axis accelerometer sensor micro-mechanical create, rate gyro and magnetometer. The method described in the "body posture monocular sequence using model following multi-stage structure." Propagation algorithm based on the grid system of confidence, data-driven Markov Chain Monte Carlo techniques. The system is a realistic scenario due to the background clutter, change in the person's appearance, self-occlusion \cite{7}. The solution is to solve the various problems associated with the data, including automatic initialization, between the self and the occlusal surface. The system is due to the attitude adequate reasoning and longer processing times resulting position error is small, and therefore not suitable for real-time applications.  The method described in the "three-dimensional body of particle filtered smooth tracking research".  Annealed particle filter, particle filter, for decomposing state level hidden Markov model technology \cite{8}. In this system, to achieve a smooth reasoning technology, occlusion and differential segmentation, hierarchical hidden Markov models and results tracking is not accurate. The system does not improve the body smooth tracking accuracy and processing time increases.

As we know, it is a mission-critical feature extraction target classification. Some researchers obtained through image length of body bone, the roof length and height, and use them to the body classification \cite{14}. This method is based on the geometric approach, a number of geometric measurements is necessary. Due to changes in attitudes and diversity body in the image, it is difficult to meet the requirements of accuracy and fast classification \cite{22}. The major contribution of this paper, we present a solution for arbitrary 3D character deformation by investigating  rotation angle of decomposition and preserving the mesh topology structure.  In computer graphics,  skeleton extraction and skeleton-driven animation is an active areas and gains increasing interests from researchers. The accuracy is critical for realistic animation and related applications. There have been extensive studies on skeleton based 3D deformation. However for the scenarios of large angle rotation of different body parts, it has been relatively less addressed by the state-of-the-art, which often yield unsatisfactory results. Besides 3D animation problems, we also notice for many 3D skeleton detection or tracking applications from a video or depth streams, large angle rotation is also a critical factor in the regression accuracy and robustness \cite{16}\cite{17}. We introduced a distortion metric function to quantify the surface curviness before and after deformation, which is a major clue for large angle rotation detection. The intensive experimental results show that our method is suitable for 3D modeling, animation,  skeleton based tracking applications. In light of the above study, the researchers conducted a motivation and different methods of gender identification and human recognition of different video database. Therefore, they decided to use an optical motion data (3D motion data) to identify the body. Since the motion capture data with the explosive growth in the past few decades, more and more interest in understanding have been developed in the field of human motion analysis and synthesis. Thus, the movement of high-quality and new applications emerge day by day basis to capture demand data or methods ubiquitous on a variety of consumer electronic devices. Select the coordinate point a number of three-dimensional motion data, select the body joint function. The function of each of the joint coordinate data of the first specification, which is calculated from the data. These functions will be used as a data point. Then the cubic Bezier curve is interpolated. The coordinates are used to calculate the sampling variance and pure-related (joint) between the articulation by using the proportion of variance. Human identification by matching the threshold value is stored in the combined value of the ratio of the combined database variance threshold to achieve.

\begin{figure}
    \includegraphics[width=8cm]{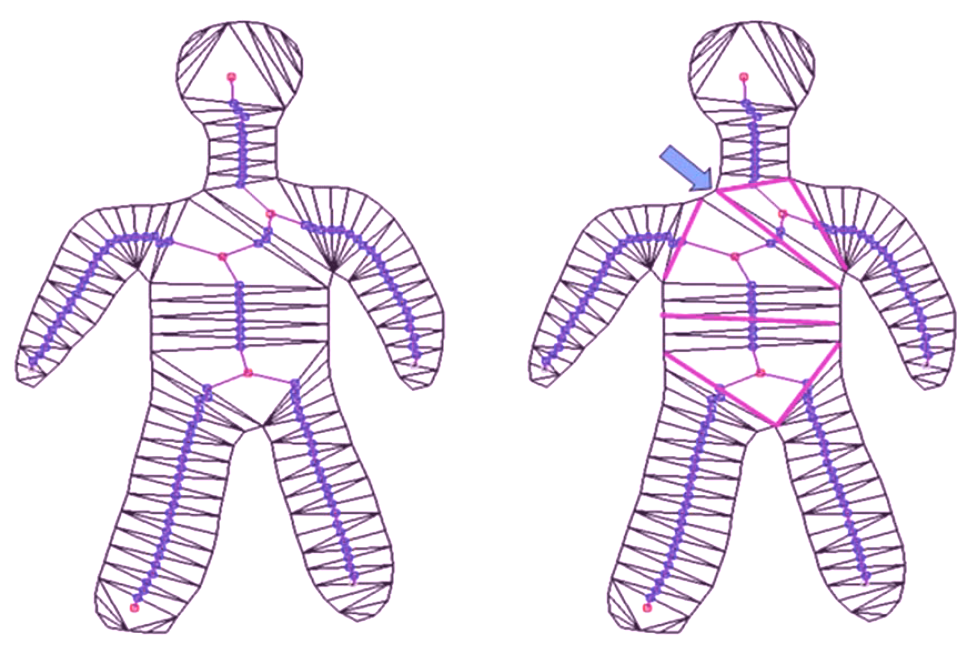}
    \caption{\label{fig 1:} Skeleton extraction results by using the angle factor}
\end{figure}

\section{Initial Skeleton Detection from 3D Mesh}
The deviation from the standard method of the Programme MLS optimize skeletal structure, rather than directly on the spot in the model. The idea is similar to the method according to the distance measured above ground, but here, the bones and joints of the skeleton of these measurements. Changes are propagated to the backbone using standard linear mixed skin mesh. The advantage of using skeleton is twofold. First, the skeleton is a good model representation of the overall model, which means that it is almost the model currently on the surface of immune small details. Secondly, in addition to a method of controlling the insertion point in trying to use, it can change the method of bone along joints and distance affect the behavior, so that part of the "rigidity" of the people more than others \cite{9}.

As described above, simulate human gesture data is an open question. In contrast, the attitude data we get the ground truth using tag-based motion capture of real human actors. The human body is a huge gesture capabilities. Joint modeling, the number of possible posture is exponential in the number of articulated joints. Therefore, we can not record all possible postures. However, there is hope. Our algorithm, based on a sliding window of forest decisions, is designed to look at a local neighborhood of pixels. Look through the windows of local, our factors combined constitute the entire body, the local position, it can be expected to summarize some of the forest unseen posture. In practice, even limited corpus, for example, each limb motion capture data are respectively constituted by broad enough. In addition, we do not need motion capture and record changes in rotation about an axis perpendicular to the mirror around, site location, body shape and size, or position of the camera, all of which can be simulated.

Our space method is to use the deformation affects a specific deformable model is only composed of a set of discrete ${Q_i}$ right. Therefore, any model will be immersed in a three-dimensional deformation of the same, irrespective of its shape. Unfortunately, in many important applications - character animation, special - required modification must consider the characteristics of the model, such as the structure and geometry of the overall shape, bone joint position. As illustration, unnatural deformation comparison shown in  Figure 1, which is a more reasonable modification. In order to reduce the negative effects of this first attempt more "form of consciousness" Euclidean metric instead of MLS to develop metrics. Figure 1 (c) illustrate MLS plans to use to measure metrics that get deformation, on the model surface distance between any two points are connected to these two points can be painted on the surface in the shortest length measurement curve. Although this idea solves some problems, it can not cope with some of the undesirable shrink / stretch effect. Most of these problems are overcome hybrid scheme combining the deformation space as possible as a rigid skeleton and animation techniques, described below. Our goal is to place a small set of appropriate control point, in order to obtain reasonable model constitution \cite{10}\cite{12}\cite{18}.

\begin{figure}
    \includegraphics[width=9cm]{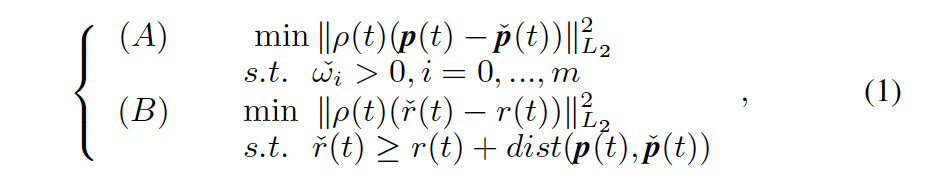}
\end{figure}

The program is divided into two stages: set (manipulated) stage, once, and deformation stage, performed once for each desired posture. In short, the first phase of manipulation to extract an approximate skeleton from the model, that is, a tree structure consisting of the joints and bones (Fig. 4) (Fig.). Next, each with a mesh vertices a backbone connection (Fig. 4 (b)). Finally, the relative influence of each joint weight distribution of a process established in any given mesh vertices (Fig. 4) (Fig.) Deformation. In the modification phase, the user first defines a set (red dots) in FIG. 4 a set of control points control point (FIG) on the. These, in turn, interact moved to the target position {Qi} This will result in the joint after being modified, MLS program described in section 3, but with the length of the measurement path instead of using the Euclidean distance norm skeleton (Fig. 4 ( e)). Each node j, the centroid {p * J} and {Q * J} and rotation vector {UJ} cache. We observed that, unlike skeletonbased character animation, our plan does not guarantee that every bone will also be critical. Finally, a hybrid linear conversion is calculated for each mesh vertices using cached values ??(Figure 4) (Figure)).

\section{Deformation Mapping Generation}
3D surface scanning is represented as a whole polygon mesh. Mesh vertices are usually defined in an arbitrary Cartesian coordinate system. Vertices must be converted to their main direction initially. Principal component analysis is a variation model to identify their weight method from the raw data (13). It has been used to find the direction of a cloud of data points is the most stretched. The coordinates of the vertices of the mesh can be converted to their geometric center of rotation of the main direction. Here, we define the direction of the point as the direction, and the direction of the point as the direction extending at least a point. Axis is the body's height, width and thickness direction, respectively.

The sliced vertically along the central axis X-ray, Y and Z are parallel to the XY, XZ plane had split point using the body due to the uncertainty and the main directions of the geometric center. A slip dislocation and the main directions of the geometric center, along the body will give significantly different pieces. Horizontal slices elaborate. Definition and center pieces evenly distributed along the x-axis, from head to foot. In each piece of light emission from their centers, and calculate the point of intersection between the surface light \cite{11}.

\begin{figure}
    \includegraphics[width=9cm]{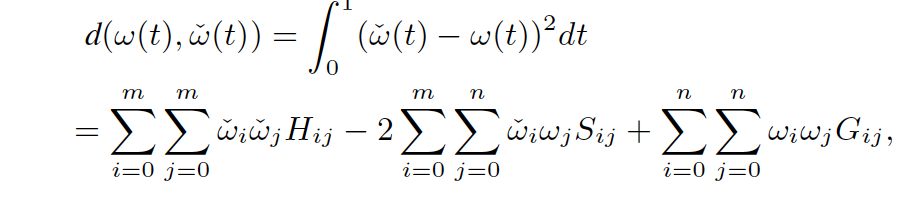}
\end{figure}

Three options are available when we slice scan the body surface. We can collect intersection nearest the center of each ray, the nearest intersection formed only of a set, or we can collect each intersection of each of the light to produce a set of all points. All intersections (option 2) settings include the nearest intersection (option 1), in addition to the arms and legs on both sides of the intersection. The results compare the two options, we can be subdivided into four limbs and trunk (Fig. 4). Calculating unit also has two steps. The first step is to extract information about the joint (shoulder, hip, knee and ankle) function, and the second step is to use these features of data points. Cubic Bezier curve by curve through the data points. Calculate the variance of each coordinate of the curve, and then calculate the average variance. Finally, the calculation unit of output as the input identification unit used. It contains two main things a detector, and the second is the database. Detector using the "ID" and compares it with the value stored identity by comparing it with "a database" are compared, and the results are given recognition results.

Third selection sections are considered unique and lead to effective segmentation. Cross light and observe the behavior of a closed grid between, we implemented a property point of intersection (Figure 5). Since the closure of the grid, there is an odd intersection (Fig. 5), when the center of the radiation within the grid, and there is even a point of intersection (Fig. 5), when the center ray is outside the grid. After launching a ray of light in a main direction, the number of the number. Even when there is an intersection, the center sections located outside the body. In Figure 5, this sequence of intersections, and form a new slice center. The new radiation is emitted from the new center of the sheet, only the new center for the collection of the closest intersection. When there is an odd number of intersection closest to the intersection of the center of the slice is discarded, for example, in Figure 5. Centre (OIN) the rest of the order of intersection (BC Fig. 5) may form a new center piece and includes the original film. Only the closest intersection of the new one.

\section{Large Angle Rotation Detection}
In the first step, we use the short space-time block train all three ways personal classification. For video data depth, we have trained a convolution network and consists of a multi-layer perceptron, we developed a constitution constitutes a descriptor, in turn, train other Multilayer Perceptron. We output a speech recognition system as a "bag, then." The specific method of processing each channel described in the following sections. Three cases, each of the space-time block and each form has been distributed in the $n + 1$ corresponding class of neuronal activation (data visualization) or class frequency (audio channel). These distributions in the longer sequence, and then connect to a recurrent neural network classifier. In all tags dynamic configuration of the testing phase, each dynamic posture $I \in [1. .. I]$ involved in the classification process is repeated as a continuous overlapping length L2 sequence $A \in [1. ..N]$ (Or less, at the border), every time a specified distribution of neurons activated $\Omega, R, N$ class $[0. .. (Nitrogen is an Index)$. Per Dynamic having an average distribution, and is calculated as follows
\begin{figure}
    \includegraphics[width=9cm]{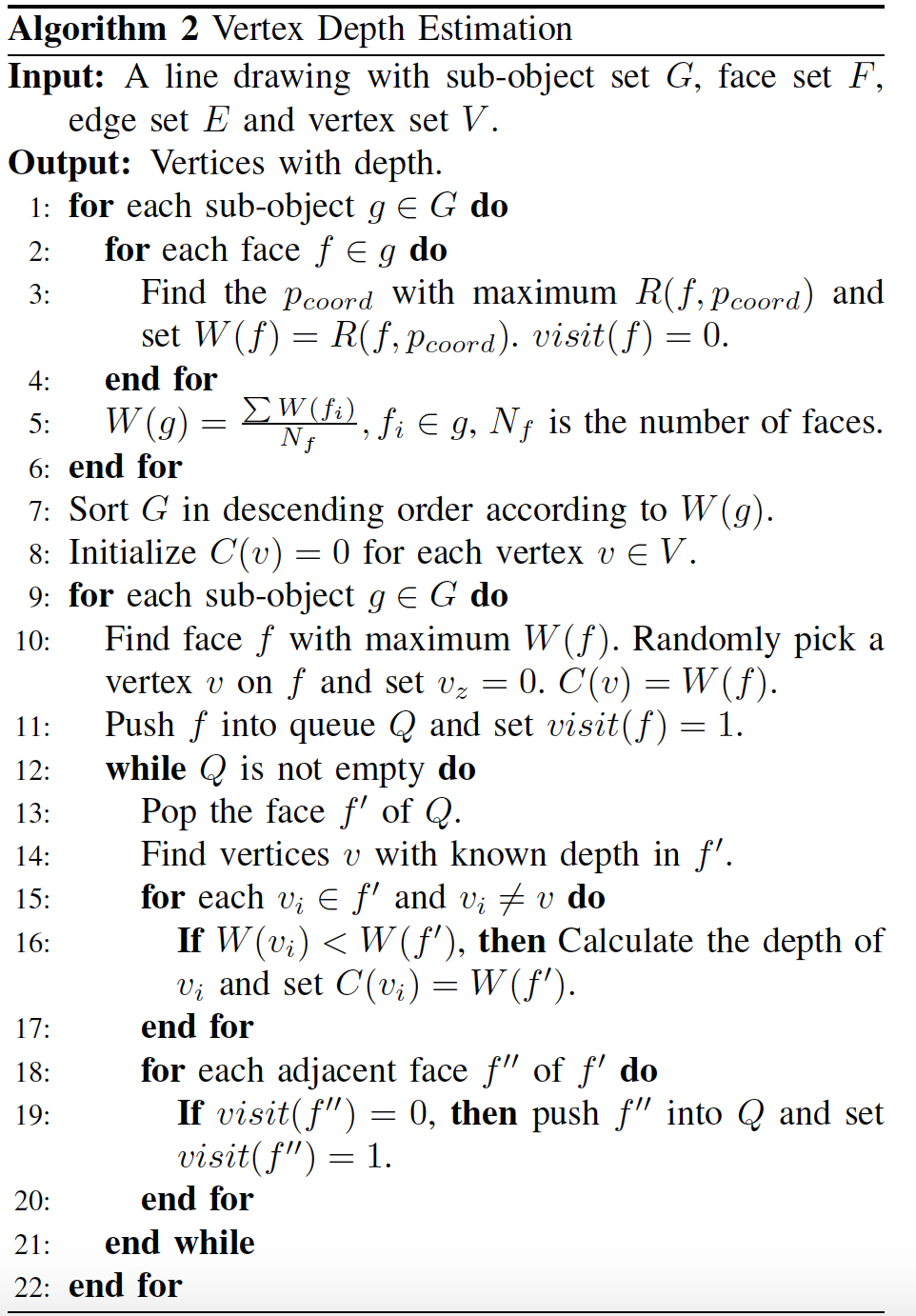}
    \caption{\label{fig 1:} Algorithm for vertex based skeleton extraction}
\end{figure}

\begin{figure}
    \includegraphics[width=9cm]{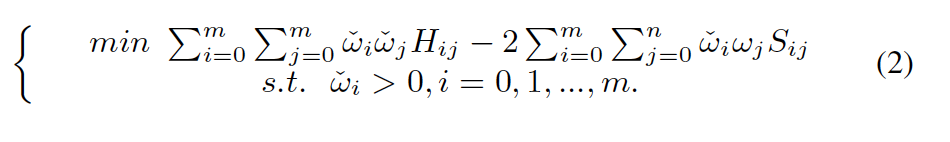}
\end{figure}
\begin{figure}
    \includegraphics[width=9cm]{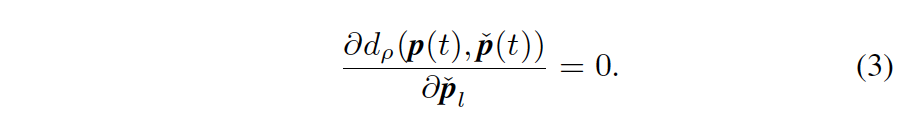}
\end{figure}

Due to the built-in smoothing parameter, skeleton track is often inertia. Thus, rapid movement of the joint position is usually detect inaccurate. Smoothing is reduced, the introduction of additional noise and jitter. To compensate for these effects, we correct position of the hand joints, minimizing the square root of the difference between the corresponding inter-block of each of the dynamic configuration. As a pre-processing step, we subtract the background from a simple threshold for each frame in the depth of the shaft, and apply the local contrast normalized to zero mean and unit variance than the local neighborhood. Finally, we use the extracted convolution network \cite{13} consisting of two layers of hyperbolic tangent activation code and 2 sub-sampling layer supervise training block (ConvNet Figure 2). Short 3D space-time block convolution (dynamic posture) on the implementation of the first layer, followed by Max together in space and time dimensions. Two-dimensional convolution and space for the implementation of the second largest pool layer. The output of the fourth layer is fully connected multilayer perceptron (MLP).

\begin{figure}
    \includegraphics[width=9cm]{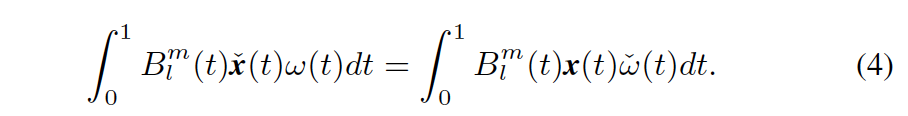}
\end{figure}
\begin{figure}
    \includegraphics[width=9cm]{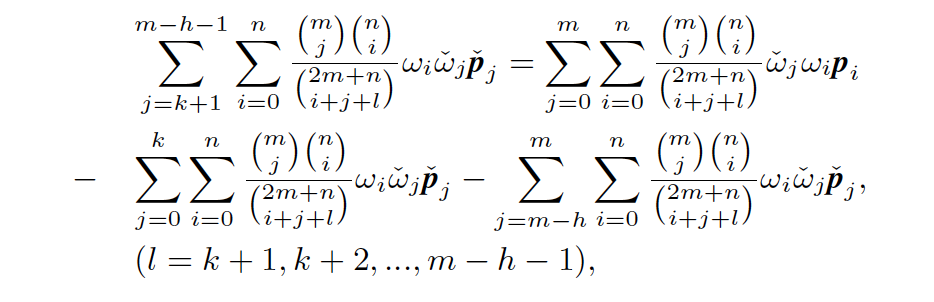}
\end{figure}
\begin{figure}
    \includegraphics[width=9cm]{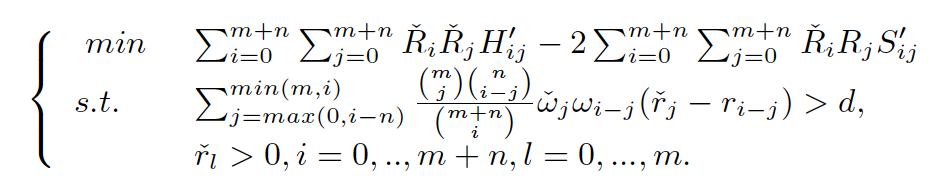}
\end{figure}

As more samples, no doubt, will reconstruct the shape close approximation function. However, there are some small planar shape given to a non-optimal sample point near L1F and L2F with shape parameters, as shown in Figure 3 (b) and 3 (C). Instead, l3F produce good shape, because it is a derivative of the first, shown in Figure 3 (D). In other words, l3f consider approaching local shape neighborhood given sample point in its. Replication is a statistical technique, which is used to minimize the extraneous variation in an experiment. While performing the experiment we tried to have its several replicates (subject walks couple of times) in order to increase precision of estimates and the cubic Bezier curve is calculated for each walk of the walking person. Let a person walk k times, and he generates the k motion curves. Using the definition of Eq. (6). These curves can be seen in Fig. 7. The normal variation of each coordinate of the each curve is calculated; and then the average variable values of the curve is computed.

\section{Experiments}
In this section, to validate our approach, we first compared with other quasi-interpolation algorithm in two approximate function, and then use our scattered point method and the traditional method of : more than three-level RBF interpolation (MRBF) , a unified multi-level partition (MPU), and screened Poisson reconstruction (shielded Poisson). Finally, a noisy data is used to show the ability of our approach to dealing with noise points. Time performance records in the PC with two Intel (R) Core (TM) i7-3770 CPU 3.4 GHz and 8 GB of RAM. It is difficult to compare our approach with other recent methods, because the input model to be segmented. Figure 3 illustrates our qualitative advantage and finished skeleton of the original spindle-aligned bounding box algorithm. In general, our method generates less than half a delicate bones dense model (Fig. 5,3,7). Therefore, if our method is used in conjunction with a rapid decomposition method, it is a very efficient overall process. Figure 5 shows the similarity Poser software and use our method to extract the skeleton generated between the default skeleton.

Assessed using statistical techniques and cubic Bezier curve method for people to recognize the performance of our proposed method, we conducted a series of standard CMU motion capture database instance [2] go. We have used the example of 37 pedestrian movement. Figure 8 shows from each of these individuals and the optimization curve values calculation formula (18). Figure 9 illustrates the results of a subject in a single walk in five sports example. Themes and Y-axis represents the view of an X-axis represents 8 optimization curve values. Figure 9 X-axis is the same, but the Y axis represents the value of a typical signature. In order to obtain satisfactory results, we have tested the remaining six examples, walk 32 different people and get a reliable accuracy. Its graphics and confusion matrix shown in FIG. 9 and 11 can be seen. Figure 10 illustrates different from figs. 8 and 9, because it contains two different topics, on behalf of walking examples
In order to evaluate the performance of classification, recognition accuracy of the test data is a major interest. We compare our results with the other two recognition functions, which are methods wherein the direction and orientation. Recognition accuracy calculation compared with the 8-direction features of the proposed method it seems to bring more computational cost, due to the combination of these two features together. In practice, we use the same pre-treatment step and direction vector extraction process two functions in our recognition system, so time costs can be reduced. We compare the consumption of time (in milliseconds) step and the recognition step of extracting the feature.

\begin{figure}
    \includegraphics[width=9cm]{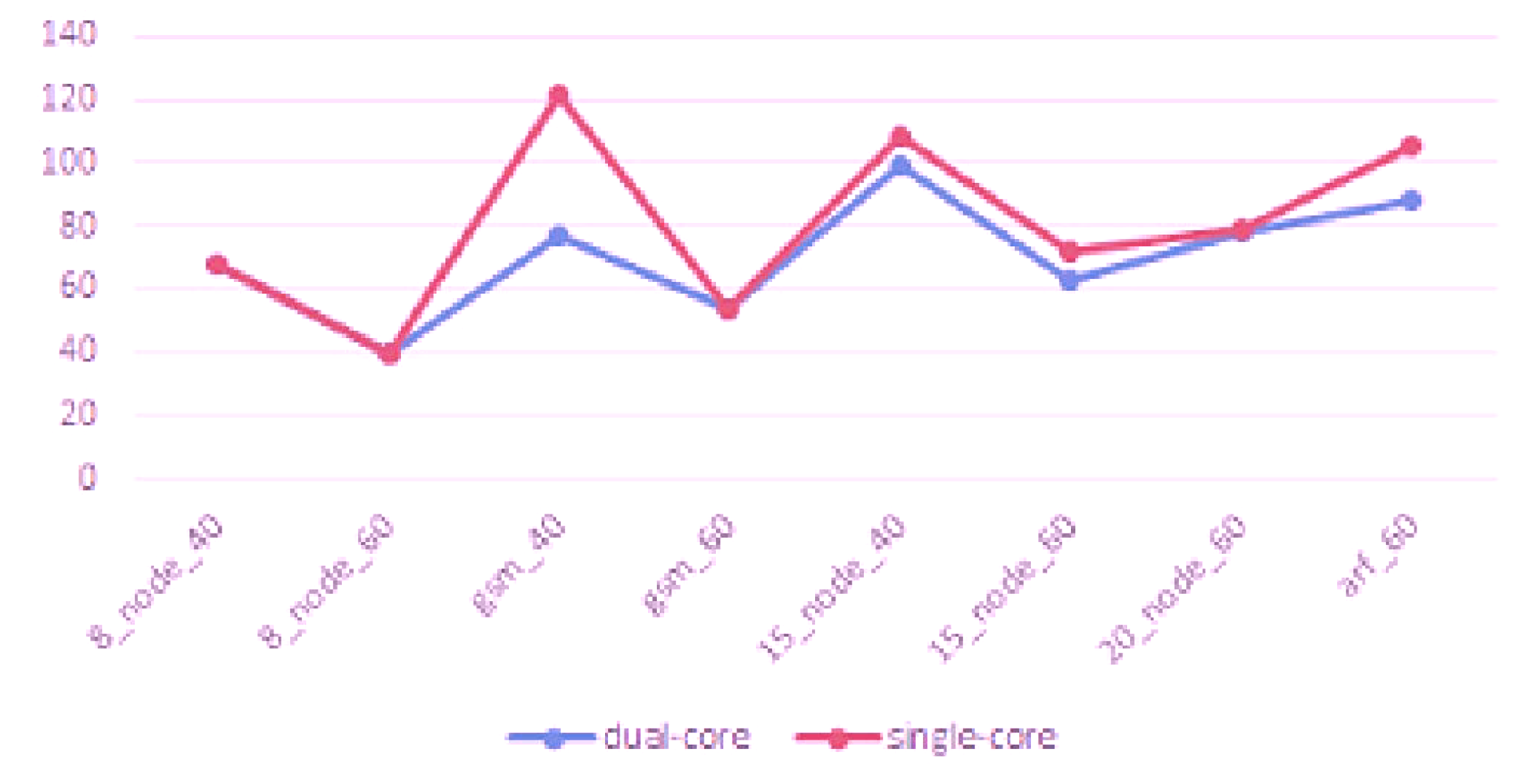}
    \caption{\label{fig 1:} Estimation error analysis }
\end{figure}

However, the total processing time of our skin law of real-time animation is very high due to the need to calculate a new point O (N1, N2 ии N3) time to complete (N1 and N2 is the point of substitution and movement and its parent components setting the base number between N3), triangulation algorithm is O (n 2) complexity, where n is the manual points. For example, Table 2 shows our rigid skinning animated characters of a human knee, from the initial movement of the extreme angle and by the time the performance of its parent component 891 vertices and 1000, respectively. Barcode complex scenes being fuzzy, low resolution. Most of these bar code module width of 1-2 pixels. In this resolution, the exact width of each column is difficult to obtain. R. Shams and P. Sadeghi proposed by the pixel close to the bar [9] boundaries to achieve sub-pixel accuracy. We use image interpolation algorithm to improve the resolution of the bar code, and by two values, rebuilding a binary image of the bar code, which can get the exact width of each bar. First, we each two adjacent linear midpoint of the projected curve. Space corresponding to the midpoint 255 of the extension, and the corresponding bar 0. The midpoint between two adjacent pixels extension line. this is,

\begin{figure}
    \includegraphics[width=9cm]{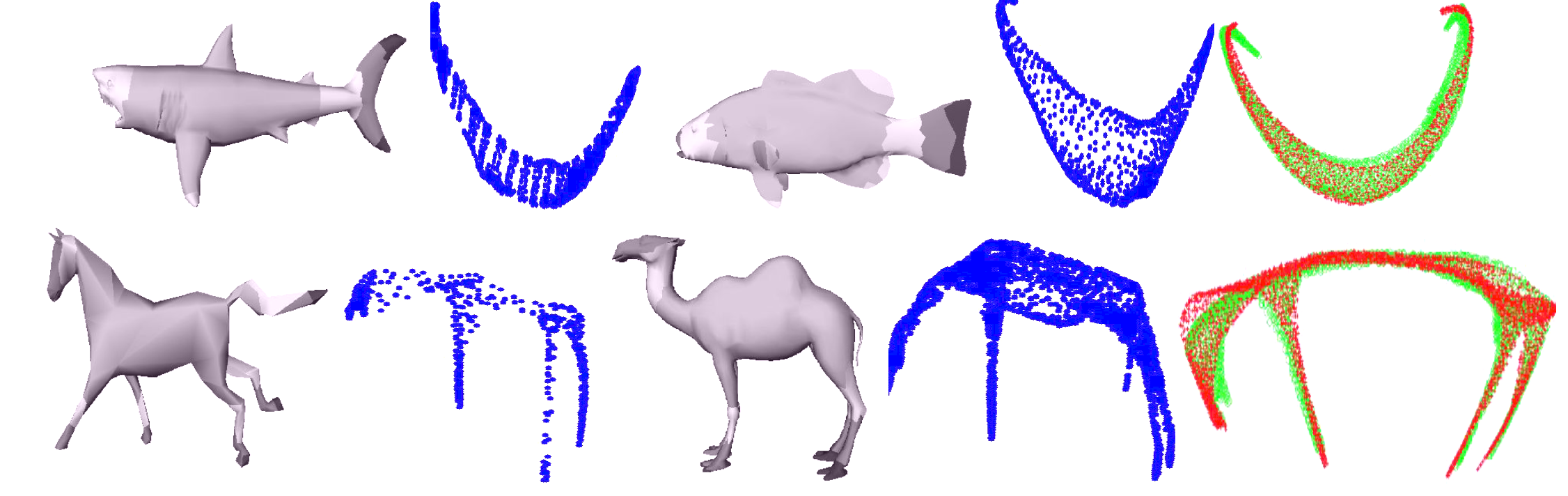}
\end{figure}

\begin{figure}
    \includegraphics[width=9cm]{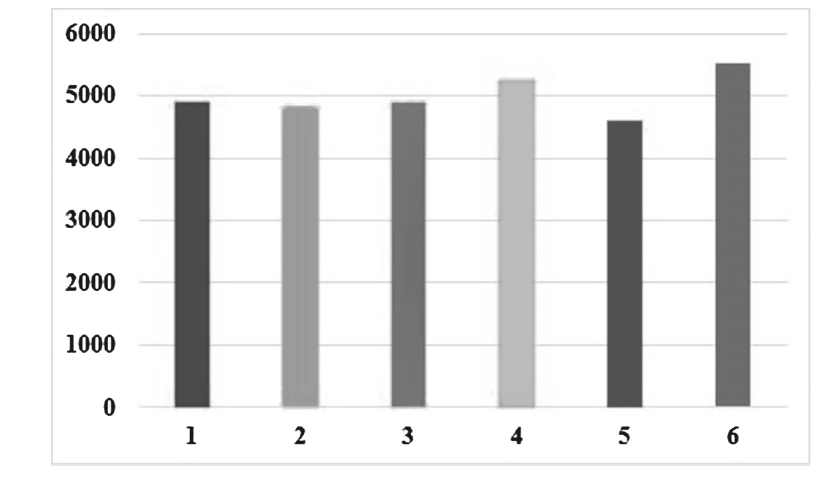}
    \caption{\label{fig 2:} Error Analysis by comparing with  other techniques }
\end{figure}

Quantitatively, given the state of the point estimate of the efficiency measured at a distance of more and better target the real target position represented by two finite sets and their respective estimated two finite sets. Figure 4 shows the GM-PHD filter and OSPA tracking performance metrics improved GM-PHD filter. On improving GM-PHD filter, Figure 4 shows that, OSPA error metric except in a few small steps, the position estimate of the true position of the target does not match when. However, OSPA error metric is quite high position estimate GM-PHD filter. Taking into account this result, it is compared to GM-PHD filter, we can confirm that there is not much risk, improve GM-PHD filter is most effective when the target is turned off. In this section, we experimentally tested the EEF in the reconstruction of shredded branches and combined with the performance of algorithms [12]. Specifically, the two methods, EEF HM HM is not evaluated. In addition, previous work proposes two methods, namely differential edge pixels [1] and the method of using the weighted difference (w-difference) edge pixels [1], but also for comparative evaluation. Crop 11 О 19 reconstruction accuracy are shown in Table 1. As shown in Table 1, significant reconstruction based on good performance in all of this work presents two methods of EEF, with HM method is superior to other methods. Specifically, for example U002 HM rose nearly 20 percent, than the other three methods accuracy.

\begin{figure}
    \includegraphics[width=9cm]{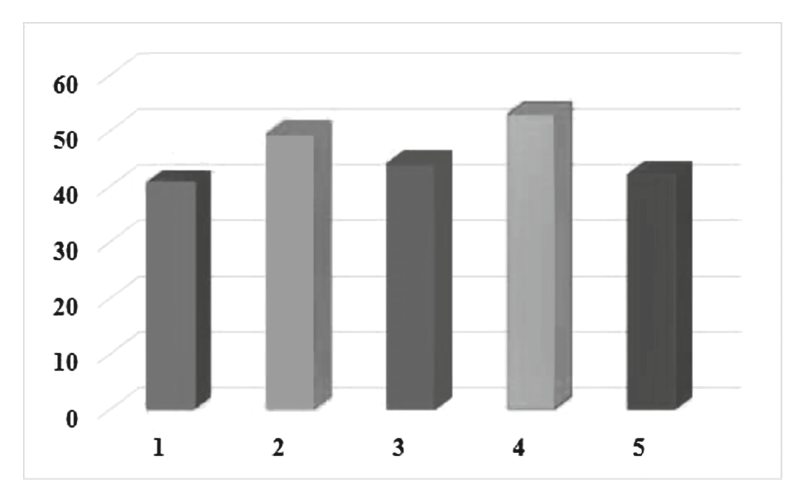}
    \caption{\label{fig 3:} Histogram of skeleton detection rate on large angle of joints }
\end{figure}
It shows from three sets of test data extracted from the skeleton, our approach and the results of the corresponding cluster initialization. Although the steps in the initialization overestimated cluster generation, our method was successfully trim excess bone, especially dance and manual models. Figure 9 demonstrates the robustness and effectiveness of our bones trim for processing a different number of clusters from the initialization step. Our method can produce a consistent structure similar to the final skeleton torso and legs, by trimming the excess bone. Note that the slight differences only in certain highly deformed areas, such as the tail and feet. In the X and Y axes on behalf of the signature value. The obtained results are determined and reported accurately accuracy rate of 100 percent, according to our data set of human recognition. The results are described in the table. 1. It describes the different areas. The first column shows the number of subjects, and the second column name of the data subjects in CMU database, and the third column shows the number of sub-themes go, the accuracy of the final column shows the percentage of body identification. We have shown that the results of this approach, as well as other existing methods and comparison tables. 2.
\begin{figure}
    \includegraphics[width=9cm]{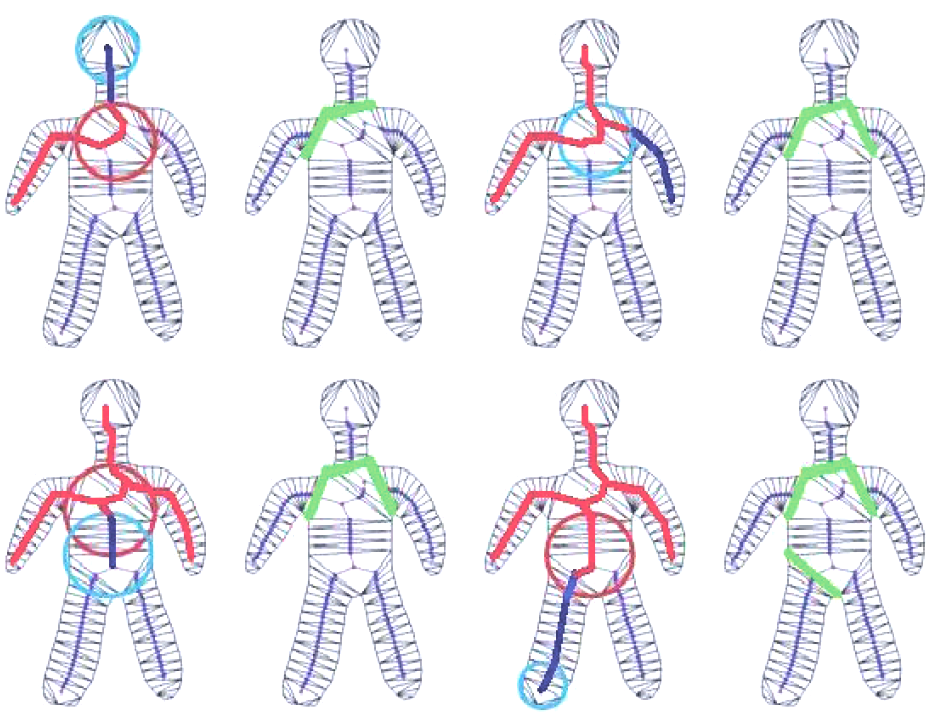}
    \caption{\label{fig 1:} Arbitrary 3D mesh structure estimation}
\end{figure}

\section{Conclusion}
In this paper, we propose a skeleton extortion algorithm based on topology rules, depth estimation algorithm, based on the depth reconstruction algorithm-level optimization. Compared with the previous face classification method, our method has strong results in non-transition stitched object. In the depth of the recovery phase, our depth estimation method greatly accelerate the reconstruction process and hierarchical optimization method based on two new laws, focusing on the transition arc and sew edges effectively avoid the local optimum. Please note that our method can not handle the line of the Liberal curved objects such as spheres and the human body. Users need to be adjusted when the intersection is three primary coordinate system, the local coordinate system of our algorithm is not suitable for the object.
\bibliographystyle{acmsiggraph}
\nocite{*}
\bibliography{template}
\end{document}